\documentclass[conference]{IEEEtran}
\IEEEoverridecommandlockouts
\usepackage{cite}
\usepackage{amsmath,amssymb,amsfonts}
\usepackage{algpseudocode}
\usepackage{graphicx}
\usepackage{textcomp}
\usepackage{makecell}
\usepackage[table]{xcolor}
\usepackage{hyperref}
\usepackage[inline]{enumitem}
\usepackage{csvsimple}
\usepackage{booktabs}
\usepackage{datatool}
\usepackage{array}
\usepackage{multirow}
\usepackage{lipsum}
\usepackage{afterpage}
\usepackage{tikz}
\usetikzlibrary{automata,arrows,positioning,calc}
\usepackage{caption}

\font\titlefont=ptmb7t at 35pt
\font\subtitlefont=ptmr7t at 21pt
\font\tinyfont=ptmr7t at 14pt

\def\BibTeX{{\rm B\kern-.05em{\sc i\kern-.025em b}\kern-.08em
    T\kern-.1667em\lower.7ex\hbox{E}\kern-.125emX}}

\usepackage{varwidth}
\DeclareCaptionFormat{myformat}{%
  \begin{varwidth}{\linewidth}%
    \centering
    #1#2#3%
  \end{varwidth}%
  }

\title{\titlefont{ Augmenting LLMs with Knowledge }\\ \subtitlefont{ A survey on hallucination prevention }\\ \tinyfont{ --- student project --- }}
\author{
\IEEEauthorblockN{Konstantinos Andriopoulos}
\IEEEauthorblockA{
Delft University of Technology\\
Delft, The Netherlands \\
A.Konstantinos@student.tudelft.nl}
\and
\IEEEauthorblockN{Johan Pouwelse}
\IEEEauthorblockA{
Delft University of Technology\\
Delft, The Netherlands \\
J.A.Pouwelse@tudelft.nl}
}

\begin{document}

\maketitle
\thispagestyle{plain}
\pagestyle{plain}

\begin{abstract}

Large pre-trained language models have demonstrated their proficiency in storing factual knowledge within their parameters and achieving remarkable results when fine-tuned for downstream natural language processing tasks. Nonetheless, their capacity to access and manipulate knowledge with precision remains constrained, resulting in performance disparities on knowledge-intensive tasks when compared to task-specific architectures. Additionally, the challenges of providing provenance for model decisions and maintaining up-to-date world knowledge persist as open research frontiers. To address these limitations, the integration of pre-trained models with differentiable access mechanisms to explicit non-parametric memory emerges as a promising solution. This survey delves into the realm of language models (LMs) augmented with the ability to tap into external knowledge sources, including external knowledge bases and search engines. While adhering to the standard objective of predicting missing tokens, these augmented LMs leverage diverse, possibly non-parametric external modules to augment their contextual processing capabilities, departing from the conventional language modeling paradigm. Through an exploration of current advancements in augmenting large language models with knowledge, this work concludes that this emerging research direction holds the potential to address prevalent issues in traditional LMs, such as hallucinations, un-grounded responses, and scalability challenges.

\end{abstract}

\section{Introduction} \label{intro}

Large Language Models (LLMs) have brought about remarkable advancements in Natural Language Processing (NLP) and are now integral to various widely-used products, including Copilot \cite{copilot}, Google's search engine, and more recently, Chat-GPT, a chatbot built upon GPT3 \cite{gpt3}. These models, characterized by their memorization capabilities as well as their compositional prowess, have demonstrated unprecedented performance in tasks ranging from language understanding to text generation, paving the way for more sophisticated human-computer interactions.

However, LLMs are not without their limitations. They often produce seemingly plausible yet incorrect predictions, a phenomenon known as hallucinations \cite{hallucinations}, leading to avoidable errors in various contexts. Furthermore, many of the groundbreaking capabilities of LLMs appear to scale with the model's size in terms of trainable parameters. While recent efforts have produced smaller LLMs with retained capabilities \cite{compute-optimal}, the practicality of training and maintaining large models remains a challenge, with continual learning for such models posing an ongoing research question \cite{continual}.

These limitations are rooted in a fundamental issue with LLMs: they are primarily trained for statistical language modeling, relying on a single parametric model and a relatively limited context, typically the preceding "n" tokens. Despite advancements in hardware and software, most models still employ relatively small context sizes compared to the expansive context required for accurate language modeling in all scenarios. Consequently, achieving the necessary scale to store the knowledge beyond the immediate context has become a necessity.

In response, a growing research trend has emerged, moving away from the traditional statistical language modeling paradigm. One approach addresses the limited context size of LLMs by enhancing its relevance through the incorporation of information extracted from external documents \cite{rag} \cite{fid}. By equipping language models with modules that retrieve relevant documents from databases based on the context, it becomes possible to replicate certain capabilities of larger LLMs while using fewer parameters \cite{retro} \cite{atlas}.

Moreover, in this evolving landscape, pioneering models \cite{graft-net} \cite{pullnet} that leverage structured knowledge stand out. These models leverage knowledge graphs along with a corpus of supporting documents, which can be jointly processed by Graph Convolutional Neural Networks (CNNs). By harnessing graph-based representations, these structured-knowledge augmented models excel in generating precise responses to open-domain questions. This innovative use of structured knowledge marks a significant advancement in enhancing language models, demonstrating the diverse strategies researchers are adopting to address the limitations of contemporary LLMs.

It is worth noting that such approaches transform the resulting models into a non-parametric ones, as they can now effectively query external data sources.

Another strategy involves enabling LLMs to leverage external tools \cite{lamda}, such as search engines \cite{internet-augmented} \cite{seeker} \cite{lamda}, allowing them to augment the current context with crucial missing information not contained within the model's weights. Although most of these efforts aim to address individual shortcomings of LLMs, it is evident that a more comprehensive integration of knowledge tools has the potential to significantly enhance the capabilities of these models.

In light of these recent developments in NLP, there is a pressing need for a comprehensive taxonomy of augmented language models and clear definitions of the technical terminology used, which sometimes carry varying interpretations and intentions.

\section{Background} \label{background}

As we delve into the intricacies of augmenting Large Language Models (LLMs) with external knowledge, it is imperative to establish a foundational understanding of the key concepts that underpin this transformative field. Knowledge augmentation strategies, such as harnessing knowledge graphs, employing beam search techniques, leveraging triplestore databases, and integrating sequence-to-sequence models, form the bedrock upon which advanced language models now stand. In this section, we embark on a comprehensive exploration of these pivotal concepts, unraveling their significance, methodologies, and interconnectedness. By elucidating these fundamental building blocks, we pave the way for a profound understanding of how contemporary LLMs harness external knowledge to achieve unprecedented linguistic feats.

\subsection{Generative Language Models}

Generative language models are trained to produce new text, given an input sequence of tokens. They are able to perform this by learning the statistical relationships between words and phrases in a large corpus of text. When given a prompt, a generative model will try to produce text that is consistent with the statistical patterns it has learned.

Some of the most popular generative models in natural language processing include autoregressive models \cite{autoregressive}, variational autoencoders (VAEs) \cite{vaes} and Generative adversarial networks (GANs) \cite{gans}. In this literature survey, we will mostly explore Transformers, autoregressive models, along with another type of generative language models, sequence-to-sequence models.

\subsection{Autoregressive Models}

An autoregressive model \cite{autoregressive} is a type of neural network used for generating sequences of data, where each element in the sequence is predicted one at a time based on the previously generated elements. In other words, the model generates data by conditioning its predictions on the data it has generated so far. Autoregressive models are typically used for tasks like text generation, time series forecasting, and speech synthesis.

One of the most well-known autoregressive models in NLP is the GPT (Generative Pre-trained Transformer) series, such as GPT-2 \cite{gpt2} and GPT-3 \cite{gpt3}. These models generate text by predicting the next word in a sentence based on the preceding words. They use self-attention \cite{transformer} mechanisms to capture dependencies between words at different positions in the sequence, making them capable of generating coherent and contextually relevant text.

\subsection{Sequence-to-sequence Models}

A sequence-to-sequence (seq2seq) model \cite{seq2seq} predicts the probability of a token being the next token in a given sequence of words.

It consists of an encoder and a decoder. The encoder reads the input sequence, one step at a time and produces a fixed-dimensional vector representation of the entire sequence. This vector is called a \textit{context vector} and it is a representation of all the meaningful information of the input sequence. The context vector is then passed to the decoder, which generates an output sequence.

Sequence-to-sequence models are typically trained using a maximum likelihood objective, which means that they are trained to produce the output sequence that is \textbf{most likely} to follow the input sequence. In summary, seq2seq models are designed for tasks involving the transformation of one sequence into another, often with different lengths and structures. They are typically applied to tasks such as: machine translation, text summarization, and question-answering, where the relationship between the input and output sequences is not purely linear or where the lengths of input and output sequences can vary significantly.

From this point and onwards, we will refer to sequence-to-sequence models as just seq2seq.

\subsection{Transformers}

The Transformer architecture \cite{transformer} marked a groundbreaking advancement in the field of NLP. Since its inception, Transformers have become the backbone of various state-of-the-art language models, underpinning many of the recent developments in the realm of augmented language models.

At its core, the Transformer architecture revolutionized sequence-to-sequence modeling through the introduction of the attention mechanism. Unlike earlier recurrent neural networks (RNNs) \cite{rnn1} \cite{rnn2} and convolutional neural networks (CNNs) \cite{cnn}, Transformers rely on self-attention mechanisms to capture dependencies between elements in a sequence, making them highly parallelizable and efficient for processing long-range dependencies.

The architecture consists of two main components: the encoder and the decoder. The encoder processes the input sequence, while the decoder generates the output sequence. Each component comprises multiple layers, with each layer containing a multi-head self-attention mechanism and feed-forward neural networks. These self-attention mechanisms enable Transformers to capture contextual information efficiently, making them ideal for tasks that involve understanding and generating sequences of data.

In the context of language modeling, Transformers can be adapted to function as decoder-only models. In decoder-only Transformers, the encoder component, which is used for encoding input sequences, is removed. These models retain the core Transformer architecture but focus exclusively on generating sequences of tokens, making them particularly suitable for autoregressive language modeling tasks.

Decoder-only Transformers operate in an autoregressive manner. They generate sequences one token at a time, with each token's prediction conditioned on the previously generated tokens. This autoregressive approach allows them to produce coherent and contextually relevant text. Decoder-only Transformers have been instrumental in various text generation tasks, including machine translation, text summarization, and text completion.

Since the introduction of the Transformer architecture, numerous variants and extensions have emerged, each tailored to address specific challenges in NLP. These variants include models such as BERT (Bidirectional Encoder Representations from Transformers) \cite{bert}, GPT (Generative Pre-trained Transformer) \cite{gpt2} \cite{gpt3}, and T5 (Text-to-Text Transfer Transformer) \cite{T5}, among others. Many of these models have laid the foundation for augmenting language models with external knowledge, a topic of great interest in recent NLP research.

\subsection{Beam Search}

Beam Search is a heuristic search algorithm that explores a graph, G, by expanding only the K (beam width) most promising nodes at each step. Beam Search simulates the behavior of Breadth-First Search. More specifically, it uses BFS to create a search tree. At each level of the tree, it checks all the successors of the current level and keeps only the top K ones, while pruning the others. The process repeats until K leaves are found. Beam search will return the leaf that maximizes some given score function.

In the context of NLP, when using a generative model, Beam Search is utilized to find the sequence $y = (y_1, ..., y_n)$ that is most likely to come after an input sequence $x$. In mathematic notation, the probability to maximize is:

\begin{equation}   
    \begin{split}
     p(y|x) & = p(y_n|x, y_{1...n-1}) \cdot p(y{1...n-1}|x) \\
     & = p(y_n|x, y_{1...n-1}) \cdot p(y_{n-1}|x, y_{1...n-2}) \cdot... \cdot p(y_1|x) \\
    \end{split}
\end{equation}

Instead of choosing only the output token with the highest probability each time, beam search chooses the top K tokens with the highest probability and explores the generated sequences recursively until we reach an $<EOS>$ (end-of-sequence) token. Then, it returns sequence $y$ (out of the K sequences) that maximizes $p(y|x)$.

In the following sections, we will explore some concepts that are pivotal to the understanding of state-of-the-art augmentation of LLMs.

\subsection{Text Corpus}

A text corpus, $D$ is a set of documents: ${d_1, ..., d_{|D|}}$ where each document is a sequence of words: $d_i = (w_1, ... , w_{|d_i|})$. Specifically, in the context of this paper, a document is essentially a sentence, and an article is a collection of documents.

As we will see later on in this survey, text corpora are considered an \textbf{unstructured} knowledge base and are usually organized in vector databases.

\subsection{Vector Database}

In a vector database, a document can correspond to one vector or many vectors, depending on the specific implementation of the database. A single vector captures the overall meaning of the document. This is often done by averaging the vectors of the words in the document. In other cases, a document may be represented by a vector for each word in the document. This is often done when it is important to be able to track the individual words in the document.

When a language model retrieves information from a vector database, it essentially has access to knowledge that is not stored in its parameters (weights). Therefore, a vector database is a form of \textbf{non-parametric memory} for LLMs.

\subsection{Dense Vector Index}

Indexing in a vector database is the process of organizing the vectors in the database in a way that makes it efficient to search for and retrieve similar vectors (vectors with a high inner product). This is accomplished by creating a data structure that maps each vector to a set of other vectors that are similar to it.

Maximum Inner Product Search (MIPS) is a specific type of vector search that aims to find the vector in the database with the highest inner product with a given query vector. MIPS is used in a variety of applications, such as recommendation systems, machine learning, and image retrieval.

FAISS \cite{faiss} is a popular open-source library for efficient similarity search and clustering of dense vectors. FAISS contains a variety of algorithms for MIPS, as well as other types of vector search. FAISS is used by many companies and organizations, including Google, Facebook, and Microsoft.

\subsection{Triplestore Knowledge Bases}

A Triplestore knowledge base is a database that consists of subject-predicate-object triples. An example of such a triple is: (Subject: Albert Einstein, Predicate: was born in, Object: Ulm, Germany). Triples are a great form of representing factual knowledge because they capture the nature of the relationship between a subject and an object and can be easily processed by LLMs. One can visualize this knowledge base as a graph whose vertices are the various subjects and objects (entities) and the predicates are the edges between these entities.

Each edge has a type (e.g: "was born in") that describes the kind of the relation between the connected entities. Triplestore knowledge bases with more than one types of relations are called \textbf{heterogeneous}.

Triplestores are an excellent example of what we call \textbf{structured} knowledge bases. They can be merged with unstructured knowledge bases through a set of \textbf{entity links}: $(v, d_p)$,  connecting entity $v$ with a word at position $p$, in document $d$.

\subsection{Graph Convolutional Networks}

Graph convolutional networks (GCNs) are a type of neural network that can be used to learn representations of nodes in a structured knowledge base, such as a graph. GCNs are particularly well-suited for node classification tasks, where the goal is to predict the label of each node in the graph (e.g: whether the node contains an answer to a given question or not).

GCNs work by iteratively aggregating information from the neighbors of each node. At each layer, the GCN collects the embeddings of all of a node's neighbors, averages them, and then applies a linear transformation and a nonlinear activation function. The output of this layer is then used as the input to the next layer.

The more layers the GCN has, the more multi-hop reasoning the model will be able to perform, because it will gather information from more far away neighbors. This makes GCNs well-suited for tasks where the labels of nodes depend on the labels of their neighbors, such as social network analysis and fraud detection.

Here is a high-level overview of how a GCN works for node classification:
\begin{enumerate}
    \item Initialize the embeddings of all nodes in the graph.
    \item For each node in the graph:
    \begin{enumerate}
        \item Collect the embeddings of all of the node's neighbors.
        \item Average the embeddings of the node's neighbors.
        \item Apply a linear transformation and a nonlinear activation function to the average embedding.
        \item The output of this function is the new embedding for the node.
    \end{enumerate}
    \item Repeat step 2 for a fixed number of layers.
    \item The final embedding of each node is used as the input to a classifier to predict the node's label.
\end{enumerate}

\subsection{Relational Graph Convolutional Networks}

One problem that arises when the knowledge-base graph heterogeneous is that, in that case, we want to take into consideration the type of relation that a node has with its neighbors before we average their embeddings.

A relational GCN \cite{rgcn} is similar to a regular GCN, but it uses a separate matrix for each type of relation. Therefore, when using a relational GCN, we aggregate the embeddings from all neighbors with a specific relation and we pass the averaged embedding into a separate CNN layer for each relation.

\section{Knowledge Base Augmented Generation} \label{knowledge-base}

Language models have the ability to store knowledge in their parameters. Alternatively, knowledge in the form of natural language can be offloaded completely from the LM by retrieving from an external knowledge base. Memory augmentation strategies help the language model to avoid producing non-factual information as well as reducing the number of parameters required to achieve comparable performance to significantly larger LMs. Based on their structure, knowledge bases can be either unstructured (text-based) or structured (graph-based). In this literature survey, we are going to explore work from both worlds.

\begin{figure}[t]
    \centering
    \captionsetup{format=plain}
    \includegraphics[width=0.45\textwidth]{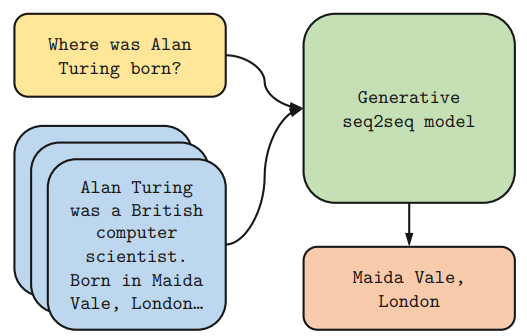}
    \caption{Overview of knowledge augmentation of language models from the paper by Izacard et al.\cite{fid}. The input query (light yellow), along with a number of retrieved relevant documents (light blue), passes through the generative seq2seq model to produce an output response.}
    \label{fig:retriever-generator}
\end{figure}

\subsection{Retrieval-Augmented Generation (RAG)} \label{sec-rag}

RAG \cite{rag} uses both parametric and non-parametric memory to generate more accurate and informative responses to an input query.

Specifically, the RAG architecture entails:
\begin{itemize}
  \item \textbf{\textit{a generator}}: a BART-large \cite{bart} sequence-to-sequence language model, pre-trained on a massive dataset of text and code (parametric memory).
  \item \textbf{\textit{a knowledge base}}: a dense vector index of the Wikipedia database (non-parametric memory). All documents in the knowledge base are also encoded as vectors using a $BERT_{BASE}$ \cite{bert} document encoder, $BERT_{d}$.
  \item \textbf{\textit{a retriever}}: a component that is responsible for retrieving the documents of the knowledge base that are most relevant to the input query. It follows the DPR (dense passage retrieval) architecture \cite{dpr} and it consists of a document encoder, $BERT_{d}$ and a query encoder, $BERT_{q}$. The retriever
  \begin{itemize}
      \item calculates the embedding of the input query, using the $BERT_{q}$ encoder.
      \item conducts \textit{Maximum Inner Product Search} (MIPS) in the indexed knowledge base to find the \textbf{K} most similar documents to the input query
  \end{itemize}
\end{itemize}

According to the authors of RAG, training and fine-tuning the parameters of the $BERT_{d}$ encoder is extremely computationally expensive, and not very effective accuracy-wise. Specifically, if they were to train the parameters of $BERT_{d}$, then for each training iteration, the embeddings of each document in the $BERT_{BASE}$ knowledge base would have to be updated as well, so that they are in-sync with the new $BERT_{d}$ encoder.

Therefore, they use a completely pre-trained $BERT_{d}$ encoder, and during the fine-tuning stage, they only fine-tune the parameters of the query encoder $BERT_{q}$.

One interesting aspect of RAG is how it implements the \textit{fusion} of knowledge from all retrieved documents to produce a final response. In both proposed versions of RAG, RAG-token and RAG-sequence, fusion is performed right after the decoder.

Specifically, RAG-token:
\begin{itemize}
  \item for each retrieved document $z$, calculates the probability for each token $y_i$ in the vocabulary to be the next token in the sequence:
  \begin{equation}
    p_{\theta}(y_i | x, z, y_{1:i-1}) \\
  \end{equation}
  \item sums the probabilities over all retrieved documents (marginalization):
  \begin{equation}
    p_{\theta}^{'}(y_i | x, y_{1:i-1}) = \sum_{z} {p_{\eta}(z | x) \cdot p_{\theta}(y_i | x, z, y_{1:i-1})} \\
  \end{equation}
  \item runs Beam Search to find the K most likely next tokens
  \item chooses the token, $y_i$ with the highest transition probability
\end{itemize}

The RAG-sequence model is quite easier to grasp. It takes into account only one retrieved document per sequence that it generates. Specifically, for each retrieved document, it conducts Beam Search to generate K sequences. Then, it simply returns the sequence with the highest probability.

\subsection{REALM \cite{realm}}

REALM was the first method that managed to pre-train jointly the retriever and the generator. The authors of REALM propose three stages of training for the given architecture:
\begin{itemize}
  \item \textbf{\textit{initialization}}
  \item \textbf{\textit{pre-training}}
  \item \textbf{\textit{fine-tuning}}
\end{itemize}

One significant challenge that REALM faced was the fact that, at the beginning of training, the query and document encoders, $Embed_{input}$ and $Embed_{doc}$ respectively contain completely random parameters. Hence, the retrieved documents, z, will likely be unrelated to the input query, x. As a result, the Generator learns to ignore the retrieved documents. Once this occurs, during training, the Retriever no longer receives a meaningful gradient and cannot improve, creating a vicious cycle that does not result in an accurate end model.

To avoid this cold-start problem, the authors warm-start (initialization) the Retriever ($Embed_{input}$ + $Embed_{doc}$) using a training objective known as the Inverse Cloze Task (ICT) \cite{ict} where, given a sentence, the model is trained to retrieve the document where that sentence came from.

In the case of the Generator, the authors warm-start it with BERT pre-training \cite{bert} and they use the uncased BERT-base model (12 layers, 768 hidden units, 12 attention heads).

After the initialization stage, the REALM proposes an  unsupervised pre-training method. During the pre-training iteration, REALM:
\begin{enumerate}
  \item randomly selects sentences from the text corpus and masks specific tokens from each sentence
  \item receives a masked query, q, as input. An example of that query would be: \textit{"The [MASK] at the top of the pyramid"}
  \item outputs its token prediction (correct answer is \textit{"pyramidion"})
  \item back-propagates through the parameters, $\theta$ of the the retriever $p_{\theta}(z|x)$, and $\phi$, of the generator $p_{\phi}(z|x)$ (joint pre-training of the models).
\end{enumerate}

During pre-training, both the $Embed_{doc}$ and the $Embed_{input}$ components of the Retriever are updated. Because the parameters of $Embed_{doc}$ are updated during pre-training, in order for the document embeddings in the Wikipedia knowledge base to stay in-sync with the updated Retriever, after each back-propagation step, REALM needs to:
\begin{enumerate}
  \item re-compute the document embeddings
  \item re-calculate the document index (in order to perform MIPS)
\end{enumerate}

This is a computationally expensive task, especially for really large databases, such as Wikipedia. Therefore, REALM was designed such that the embedding updates happen every 100 back-propagation steps, as an asynchronous process.

The supervised fine-tuning method that the authors used in order to evaluate REALM on Open-domain Question Answering (Open-QA) goes as follows:
1. they collect question-answer tuples, such as: (\textit{"What's the angle of an equilateral triangle"}, \textit{"60 degrees"}).
4. REALM receives the question as input.
5. it outputs its prediction.
6. similar to the pre-training phase, REALM back-propagates through the parameters of the the retriever $p_{\theta}(z|x)$, and $\phi$, of the generator $p_{\phi}(z|x)$, but this time $Embed_{doc}$ stays untouched. Therefore, fine-tuning is much less computationally expensive.

\subsection{Fusion in Decoder (FiD)} \label{sec-fid}

FiD \cite{fid} employs a similar but quite simpler idea to RAG. Their main difference, however, lies in the way they perform the fusion of the retrieved knowledge.

Similar to RAG, in FiD, we have two main models:
\begin{itemize}
  \item \textbf{\textit{the retriever}} which has access to a $BERT_{BASE}$ where documents are represented as dense vectors and retrieves the most relevant documents by running \textit{Maximum Inner Product Search} (MIPS) using the FAISS library \cite{faiss}
  \item \textbf{\textit{the generator}} which is a sequence-to-sequence model that receives the input query concatenated with a retrieved passage and is trained to produce an answer. For their experiments, they used a pre-trained T5 \cite{T5} seq2seq model.
\end{itemize}

In FiD, fusion of the knowledge in the retrieved documents is performed right before the decoder. Specifically, similar to RAG, they concatenate the input query with each retrieved passage and they separately feed each concatenation to the encoder (in parallel). However, after that, all the produced encoded vectors are concatenated together (fusion) and are passed a single-vector input to the decoder, which performs attention across all retrieved documents (cross-attention).

\begin{figure*}[t]
    \centering
    \captionsetup{format=plain}
    \includegraphics[width=0.95\textwidth]{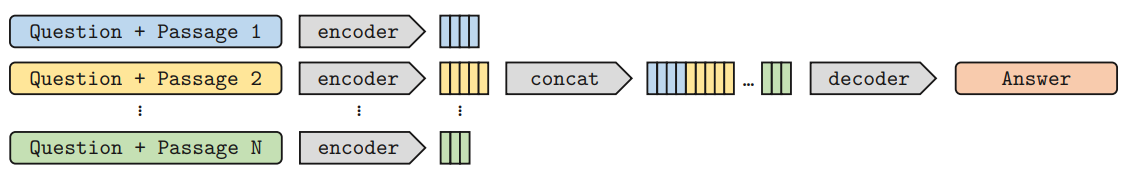}
    \caption{Overview of the Fusion-in-Decoder (FiD)\cite{fid} technique. The input question gets concatenated with each relevant passage and all concatenations get encoded in parallel. The embeddings that are produced are concatenated together (fusion) and are passed as input to the decoder.}
    \label{fig:retriever-generator}
\end{figure*}

\subsection{Atlas}

Atlas \cite{atlas} is essentially the next generation of RAG and FiD, but it specializes in few-shot learning tasks. Atlas builds upon REALM \cite{realm} and proposes jointly pre-training both the retriever and the generator model, unlike RAG which uses pre-trained models and jointly trains them only during the fine-tuning stage.

When performing a task, from question answering to generating Wikipedia articles, Atlas starts by retrieving the top-k relevant documents from a large corpus of text. Then, these documents are fed to the language model, along with the query, which in turn generates the output. Both the retriever and the language model are based on pre-trained transformer networks.

Atlas, similar to FiD, follows the retriever-generator architecture:
\begin{itemize}
  \item \textbf{\textit{the retriever}} is based on the \textit{Contriever} model \cite{contrastive} which entails a $BERT_q$ and a $BERT_d$ encoder and returns the K most relevant documents based on their similarity with the query.
  \item \textbf{\textit{the generator}} uses a T5 seq2seq model \cite{T5} and applies the FiD technique that processes each document separately in the encoder and concatenates the embeddings before they enter the decoder.
\end{itemize}

Atlas, in contrast with RAG, trains both $BERT_q$ and $BERT_d$ (not only $BERT_q$). Hence, the $BERT_q$ embeddings for each document in the knowledge base need to be regularly updated so that they are in-sync with the updated $BERT_d$ encoder. This is a computationally expensive task.


\subsection{RETRO}

The creators of RETRO \cite{retro} managed to implement an augmented language model at an unprecedented scale. This work's breakthrough is that it managed to pre-train and augment a relatively small Transformer model (25×fewer parameters than GPT-3 \cite{gpt3}) with a database that is 2 trillion tokens large ($10^3$×larger than similar retrieval-augmented LLMs).

As we saw in previous work, such as RAG, REALM and Atlas, one main difficulty of augmenting LLMs with external knowledge-bases is that training the Retriever can be computationally expensive, because while the document encoder becomes better, the embeddings for each passage in the database need to be recomputed.

In this paper, they completely bypassed that challenge by using a frozen BERT retriever \cite{bert} which contains a pre-trained document encoder. Hence, in RETRO they calculate the document embeddings once, in the beginning, and do not update them again. As a result, the main bottleneck that accessing the external database entails is to retrieve the K most-relevant documents to the input query, which they implemented using the SCaNN library \cite{scann}. This is a task of sub-linear complexity, which means that we can query their 2 trillion token database in 10ms.

One main difference of RETRO with previous work is that in RETRO they don't retrieve single documents (sentences), but chunks (a retrieved sentence along with the following sentence). This enables the generator model to acquire more context around the retrieved information and produce more accurate answers.

Here is an overview of how RETRO produces an answer to an input query, q:
\begin{enumerate}
    \item it splits the input query into chunks of 4 tokens
    \item For each chunk, cq of q, RETRO: \begin{enumerate}
        \item calculates the embedding of the chunk
        \item finds the 2 nearest neighbors (most relevant documents) in its knowledge base
        \item encodes cq through the encoder
        \item encodes the 2 nearest neighbors through the encoder
        \item interleaves the encodings of the nearest neighbors with the query-chunk embeddings to perform cross-attention. Neighbors of the first chunk only affect the last token of the first chunk and the first tokens of the second chunk.
    \end{enumerate}
\end{enumerate}

Through this technique, RETRO manages to perform attention in complexity that is linear to the number of retrieved passages.

\subsection{GRAFT-Net}

GRAFT-Net \cite{graft-net} is a novel model designed for enhancing Question Answering (QA) in scenarios where there is a structured, graph-like knowledge base (triplestore) along with a substantial text corpus. GRAFT-Net leverages advancements in graph representation learning to extract answers by creating question-specific sub-graphs containing both text and knowledge-base entities and relations.

Results in a range of benchmarks demonstrate that GRAFT-Net exhibits competitive performance compared to state-of-the-art methods when tested on either structured knowledge bases or text corpora in isolation.

Graft-Net consists of the following stages:
\begin{enumerate}
    \item the question sub-graph ($G_q$) retrieval stage: This is a characteristic of early fusion, the process of combining information from the triplestore knowledge-base and text early in the model, i.e., before a graph neural network is used.
    \item the answer selection stage, where GRAFT-Net use a Graph Convolutional Network (GCN) variant \cite{kipf2017semisupervised} \cite{li2017gated} \cite{rgcn} to do binary classification (answer, not-answer) on the nodes of $G_q$.
\end{enumerate}

The question sub-graph $G_q$ essentially is a copy of the entire knowledge-base graph, in which the nodes and edges that are irrelevant to a given question, $q$, are pruned. In addition, the question sub-graph contains text documents as well, but only the ones that are likely to contain the answer to question $q$.

The retrieval of the question sub-graph, $G_q$ happens in two parallel pipelines:

\begin{enumerate}
    \item Knowledge Base Retrieval
    \item Text Retrieval
\end{enumerate}

During the knowledge base retrieval, a sub-graph of the triplestore knowledge base is retrieved. Specifically, GRAFT-Net:

\begin{enumerate}
    \item retrieves a set of seed entities, $Sq$, that are relevant to the question $q$
    \item runs the Personalized PageRank (PPR) method \cite{ppr} around these seeds to identify other entities which might be an answer to the question. During PPR, we assign weights to edges around the seed entities. Each edge weight is essentially the cosine similarity between:
    \begin{itemize}
        \item the question vector, v(q): average of all word vectors in the question
        \item the relation vector, v(r): average of all word vectors in the relation corresponding to that edge
    \end{itemize}
    \item retains the top E entities $v_1$, ..., $v_E$ by PPR score, along with any edges between them, and adds them to the question sub-graph, $G_q$
\end{enumerate}

During the text retrieval phase, GRAFT-Net retrieves documents (sentences) relevant to the question , $q$, from the Wikipedia database. The text retrieval phase entails the steps that are described below. GRAFT-Net: 

\begin{enumerate}
    \item retrieves the top 5 most relevant Wikipedia articles (collection of documents), by using a weighted bag-of-words model \cite{drqa}.
    \item populates a Lucene index \cite{lucene} (facilitates data search in a large corpus of text) with sentences from these articles, and retrieves the top ranking ones: $d_1$, ..., $d_D$.
\end{enumerate}

The final question graph $G_q$ consists of:

\begin{itemize}
    \item $V_q$: all retrieved entities and documents
    \item $E_q$: all relations between the retrieved entities and all entity links between entities and documents
\end{itemize}

Because the vertices of the graphs can be either entities or documents, the graph is considered: heterogeneous.

\subsection{PullNet \cite{pullnet}}

PullNet builds upon the advancements made by GRAFT-Net and uses the text corpus to supplement information extracted from the triplestore knowledge base in order to answer multi-hop questions. The subjects and objects in the triples contain links to relevant documents in the text corpus and PullNet uses these links to produce more factually-based answers.

Like GRAFT-Net, PullNet has an initial phase where it retrieves a question sub-graph $G_q$. However, PullNet \textbf{learns} how to construct the sub-graph, rather than using an ad-hoc subgraph-building strategy. More specifically, PullNet relies on a small set of retrieval operations, each of which expands a graph node by retrieving new information from the knowledge base or the corpus. It learns when and where to apply these “pull” operations with another graph CNN classifier. The “pull” classifier is weakly supervised, using question-answer pairs.

The end result is a learned iterative process for sub-graph construction, which begins with a small sub-graph containing only the question text and the entities which it contains, and gradually expands the sub-graph to contain information from the knowledge base and corpus that are likely to be useful. The process is especially effective for multi-hop questions.

\section{Search-Engine Augmented Generation} \label{search-engine}

Augmenting large language models with search engines represents the next step in the evolution of AI-driven natural language processing. Search engines empower models with a gateway to an expansive universe of knowledge that far surpasses what external knowledge bases can access. By harnessing the prowess of search engines, these models gain the ability to tap into the vast and ever-expanding repository of information on the World Wide Web. This dynamic access not only provides a wealth of information but also ensures that text generation remains current and up-to-date with the latest developments, a feat that external knowledge bases often struggle to achieve as they require continuous updates.

However, it is crucial to acknowledge that this newfound access to the open web through search engines carries potential risks. The information landscape of the internet is diverse, encompassing both valuable knowledge and, regrettably, harmful or malicious content. When integrated with augmented large language models, there exists the possibility of inadvertently exposing the model to inappropriate or unsafe content. This introduces concerns regarding the reliability and safety of the generated responses, as the model may unintentionally incorporate harmful information into its outputs.

As we will see in the following sections, the use of search engine-based queries has the benefit that these queries are inherently designed to be understood by humans, enhancing both the interpretability of the model's responses and its potential for continuous improvement through direct annotation or feedback. However, to harness the immense potential of this symbiotic fusion of AI-driven language models and the vast knowledge landscape facilitated by search engines, it is imperative to develop robust safeguards and mechanisms to mitigate the risks associated with accessing potentially harmful or malicious content. This will ensure that the augmentation of language models with search engines not only broadens their horizons but also maintains the integrity and safety of their outputs, ushering in a new era of responsible and informed natural language understanding and interaction.

\subsection{Internet Augmented Dialogue Generation (IADG)}

Previously described FAISS-based approaches, such as RAG (\ref{sec-rag}) and FiD (\ref{sec-fid}), can take advantage of many existing methods developed for QA and dialogue tasks, as we saw, but have several disadvantages. First, they may be difficult to update to real-time web documents. On top of that, there may be a limit to the number of documents that can be stored in local FAISS deployments. Finally, such methods will not take advantage of the high quality ranking that has been finely tuned in Internet Search engines over decades of use. Thus, the authors of this paper by Facebook AI Research consider using Internet search engines directly for knowledge retrieval.

IADG \cite{internet-augmented} consists of two main components:
\begin{itemize}
    \item a search query generator: an encoder-decoder Transformer that takes in the dialogue context as input, and generates a search query. This is given to the black-box search engine API, and N documents are returned.
    \item a FiD-style generator: an encoder-decoder model that encodes each document individually (along with the dialog context), concatenates the embeddings before they enter the encoder, and finally generates the next response.
\end{itemize}

Each of these components can be trained separately, given \textbf{supervised} data for both tasks. The query generator requires: (context, search query) pairs, and the response generator requires: (context, response) pairs.

The search engine is a black box in this system (similar to LaMDA), and could potentially be swapped out for any method. In IADG, they use the Bing Search API \cite{bing} for their experiments to generate a list of URLs for each query. Then, they use these URLs as keys to find their page content.

\subsection{SeeKeR}

SeeKeR \cite{seeker} (Search-engine $\rightarrow$ Knowledge $\rightarrow$ Response) introduces an innovative approach that employs a single language model to tackle three distinct modular tasks consecutively: searching for information, generating knowledge, and crafting a final response. In this research endeavor, SeeKeR explores a modular framework that builds upon the foundations of IADG \cite{internet-augmented} while amalgamating the most effective elements from various existing solutions.

The SeeKeR model adheres to the foundational architecture of the standard transformer \cite{transformer}, but it distinguishes itself by employing the same model in a modular fashion, iteratively for multiple tasks. Within each module, the encoder (or decoder) incorporates distinct special tokens to signal the specific module being activated. The output generated by each module is subsequently fed into the next one, along with the original context. SeeKeR comprises a trio of specialized modules, each dedicated to unique functionalities, namely:

\begin{itemize}
    \item Search module: generates a search query from the encoded input context. Subsequently, this query is channeled into the Bing Web Search API \cite{bing}, initiating a retrieval process that yields the 5 most relevant documents as outcomes.
    \item Knowledge module: utilizes the encoded input context and a pool of retrieved documents to generate meaningful responses. This response comprises one or more pertinent phrases or sentences extracted directly from the retrieved documents. Notably, the FiD \cite{fid} method is employed to encode both the context and the documents.
    \item Response module: operates on the encoded input context merged with the knowledge response and crafts a coherent and contextually relevant continuation to the input.
\end{itemize}

It is essential to highlight that the knowledge module essentially involves a "copy" mechanism, as it does not entail the creation of new tokens; rather, its complexity lies in the precise selection of the relevant knowledge to replicate.

The authors of SeeKeR consider the GPT2 transformer \cite{gpt2} as a base model, and fine-tune it to become a SeeKeR model. Therefore, they did not perform any pre-training of their own in this case. For their experiments, they considered medium, large and XL (345M, 762M and 1.5B parameters) models.

\subsection{LaMDA}

In this paper by Google, the authors of LaMDA \cite{lamda} manage to augment a language generation model with what they call a Toolset  (TS), a black-box external knowledge source. The Toolset consists of:
\begin{enumerate}
    \item a calculator
    \item a translator
    \item an information retrieval system (similar to a search engine)
\end{enumerate}

The TS takes a single string as input and outputs a list of one or more strings. Each tool in TS expects a string and returns a list of strings. For example, the information retrieval system can receive \textit{“How old is Rafael Nadal?”} as input, and output [\textit{“Rafael Nadal / Age / 35”}].

The information retrieval system is also capable of returning snippets of content from the open web, with their corresponding URLs. The TS tries an input string on all of its tools, and produces a final output list of strings by concatenating the output lists from every tool in the following order: calculator, translator, and information retrieval system. A tool will return an empty list of results if it can’t parse the input (e.g., the calculator cannot parse \textit{“How old is Rafael Nadal?”}), and therefore does not contribute to the final output list.

It is essential to note that only little information is given on how the information retrieval system works, in the LaMDA paper, apart from the fact that it entails a database, but also can provide web snippets along with their URLs.

LaMDA entails two main sub-models that follow the decoder-only Transformer architecture:

\begin{enumerate}
    \item \textbf{\textit{LaMDA-Base}}: A regular generative model that is pre-trained on a large dataset. LaMDA-Base is the first model to receive a query from the user. It then generates a response that is checked and refined by LaMDA-Research.
    \item \textbf{\textit{LaMDA-Research}}: A generative model that usually receives the output of LaMDA-Base as input and is fine-tuned to choose the recipient of its output (the TS or the user). In general, LaMDA-Research queries the TS in a loop, until it has sufficient information to generate a final response to the user.
\end{enumerate}

\begin{table*}[t]
\centering
\begin{tabular}{|c|c|c|c|c|}
\hline
\textbf{Year} & \textbf{ALM} & \textbf{Source of Knowledge} & \textbf{Retriever} & \textbf{Generator} \\
\hline
2018 & GRAFT-Net & Graph + Text & Personalized PageRank + DrQA & GCNN \\

2019 & PullNet   & Graph + Text & Pull                         & GCNN \\
2020 & RAG       & Text         & BERT                         & seq2seq \\
2020 & REALM     & Text         & BERT                         & seq2seq \\
2021 & FiD       & Text         & BERT                         & seq2seq \\
2021 & IADG      & Internet     & seq2seq + Search Engine      & Encoder-Decoder Transformer \\
2022 & LaMDA     & Internet     & Black Box Information Retrieval System & Decoder-only Transformer \\
2022 & Atlas     & Text         & Contriever                   & seq2seq \\
2022 & RETRO     & Text         & BERT                         & Encoder-Decoder Transformer \\
2022 & SeeKeR    & Text         & Encoder-Decoder Transformer  & Encoder-Decoder Transformer \\
\hline
\end{tabular}
\caption{Overview of mentioned augmented language model (ALM) architectures}
\label{table:1}
\end{table*}

\section{Limitations and Discussion} \label{limitations}

Augmented large language models grapple with a set of recurring challenges. These issues encompass occasional inconsistencies, contradictions, factual inaccuracies, potential repetition, and a limited depth of reasoning, among others \cite{recipes} \cite{training-instructions}.

Furthermore, concerns emerge regarding the generation of content imbued with toxic language and bias, especially in specific contexts and topics \cite{recipes-safety} \cite{queens}. Another noteworthy concern is the influence of internet-sourced documents on model outputs, potentially leading to the retrieval of undesirable content. Many research experiments lean on externally developed search engines, offering advantages in terms of optimization and reliability. However, building one's retrieval system, as is often the case in question-answering (QA) and language modeling (LM) research, necessitates starting from scratch.

While search engines are adept at crawling and indexing the latest news and documents, this process demands significant engineering effort and is vital for various applications. Conversely, methods in the literature using their retrieval setups often rely on fixed document databases, which become outdated over time. Additionally, search engines are designed for human interaction, using natural language queries with limited context. In contrast, machine-generated queries, as exemplified by models like RAG \cite{rag}, can potentially encode more context or adopt vector-encoded queries, albeit at the cost of human interpretability. A benefit of search engine-based queries is their human readability, offering both interpretability and the potential for improvement through direct annotation or feedback.

Language models employing augmentation address the challenge of hallucination but do not guarantee factual grounding. Instances of conflicting retrievals can lead to mixed responses. To enhance reliability, the introduction of trust mechanisms, assigning different weights to retrievals, is a potential avenue. Another concern is the generation of generic responses that may overlook the incorporated knowledge.

In this survey, we have highlighted these common challenges and limitations faced by augmented large language models, shedding light on the evolving landscape of language generation and the pressing need for innovative solutions.




\section{Conclusion} \label{conclusion}

In this literature survey, we have explored a multitude of works in which Language Models (LMs) have been enriched with external knowledge, enabling them to generate more contextually grounded and up-to-date responses. Throughout these studies, LMs have demonstrated their capacity to enhance context by incorporating relevant information, thereby fostering the production of informative answers to various questions. This augmentation often involves the integration of non-parametric modules, marking a departure from the conventional language modeling paradigm and categorizing these models as augmented language models.

However, it is essential to acknowledge certain limitations within this paradigm shift. While LMs augmented with external knowledge exhibit reduced hallucination, they do not offer an ironclad guarantee of factual grounding. Instances arise where conflicting retrievals result in mixed answers, underscoring the need for continued refinement in this domain. Moreover, the limited exploration of the interplay between reasoning augmentation and knowledge integration in current research highlights a promising avenue for future endeavors.

As we reflect on the landscape of augmented language models, it becomes evident that this field holds immense promise and excitement. It represents a vital step towards ushering in the next generation of deep learning systems that can engage in complex and meaningful human-machine interactions while minimizing the parameter footprint. The journey towards fully realizing the potential of augmented LMs is ongoing, with opportunities for further innovation and investigation awaiting those who seek to shape the future of this dynamic field.


\bibliographystyle{IEEEtran}
\bibliography{references}

\end{document}